\documentclass[conference]{IEEEtran}
\IEEEoverridecommandlockouts
\usepackage{cite}
\usepackage{amsmath,amssymb,amsfonts}
\usepackage{algorithmic}
\usepackage{graphicx}
\usepackage{textcomp}
\usepackage{xcolor}
\def\BibTeX{{\rm B\kern-.05em{\sc i\kern-.025em b}\kern-.08em
    T\kern-.1667em\lower.7ex\hbox{E}\kern-.125emX}}
    
\usepackage{url}
\usepackage[hidelinks]{hyperref}
\usepackage[small]{caption}
\usepackage{booktabs}
\usepackage{algorithm}
\usepackage{multirow}
\usepackage{array}
\usepackage{enumitem}
\usepackage{tabularx}
\usepackage[normalem]{ulem}
\usepackage{subfig}
\usepackage{wasysym}

\usepackage{array}
\newcolumntype{H}{>{\setbox0=\hbox\bgroup}c<{\egroup}@{}}

\usepackage{tikz}

\normalsize{%
\setlength\abovedisplayskip{-6pt}
\setlength\belowdisplayskip{0pt}
\setlength\abovedisplayshortskip{-6pt}
\setlength\belowdisplayshortskip{0pt}
}
\newcommand{\mymethod}{\textsf{Malcom}}
\newlength{\myspace}
\begin{document}

\author{\IEEEauthorblockN{Thai Le, Suhang Wang, and Dongwon Lee}
\IEEEauthorblockA{The Pennsylvania State University, USA\\
\{thai.le, szw494, dongwon\}@psu.edu}
}

\title{MALCOM: Generating Malicious Comments to Attack Neural Fake News Detection Models}

\maketitle

\begin{abstract}
In recent years, the proliferation of so-called ``fake news" has caused much disruptions in society and weakened the news ecosystem. Therefore, to mitigate such problems, researchers have developed state-of-the-art (SOTA) models to auto-detect fake news on social media using sophisticated data science and machine learning techniques. In this work, then, we ask ``what if adversaries attempt to attack such detection models?" and investigate related issues by (i) proposing a novel attack scenario against fake news detectors, in which adversaries can post malicious comments toward news articles to mislead SOTA fake news detectors, and (ii) developing {\mymethod}, an end-to-end adversarial comment generation framework to achieve such an attack. Through a comprehensive evaluation, we demonstrate that about 94\% and 93.5\% of the time on average {\mymethod} can successfully mislead five of the latest neural detection models to always output targeted real and fake news labels. Furthermore, {\mymethod} can also fool black box fake news detectors to always output real news labels 90\% of the time on average. We also compare our attack model with four baselines across two real-world datasets, not only on attack performance but also on generated quality, coherency, transferability, and robustness. We release the source code of {\mymethod} at \url{https://github.com/lethaiq/MALCOM}\footnote{This work was in part supported by NSF awards \#1742702, \#1820609, \#1909702, \#1915801, \#1934782, and \#IIS1909702}.
\end{abstract}


\vspace{-5pt}
\section{INTRODUCTION}

\begin{figure}[tb]
\centering
\caption{A malicious comment generated by {\mymethod} misleads a neural fake news detector to predict real news as fake.}
\label{tab:notations}
{\small
\begin{tabular}{|l|}
    \multicolumn{1}{c}{
    \hspace*{-0.75cm}
    \includegraphics[width=0.70\linewidth]{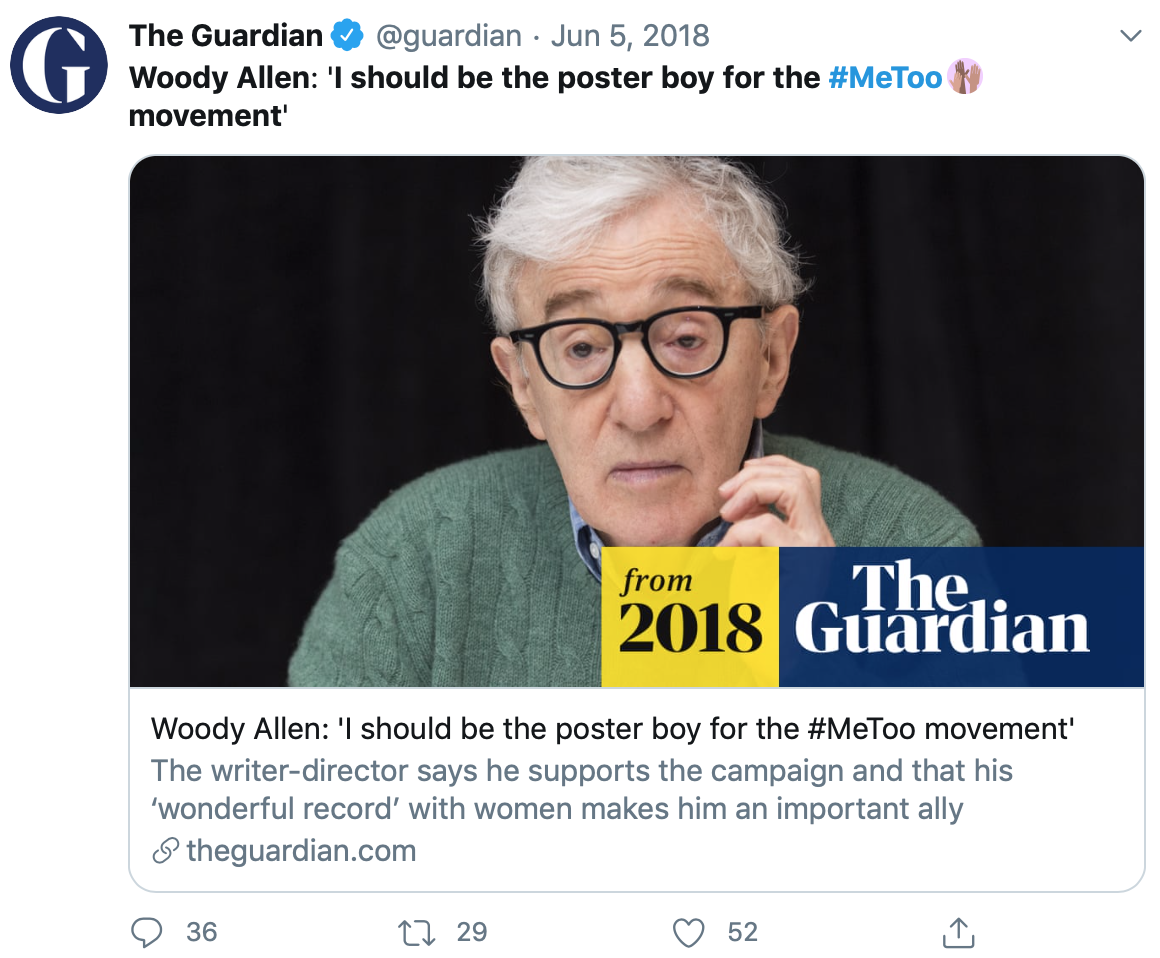}
    }\\
    \hline
    \multirow{1}{7.7cm}{\textbf{Real Comment:} admitting i’m not going to read this (...)}\\
    \hline
    \multirow{1}{7.7cm}{{\mymethod}: \textit{he’s a conservative from a few months ago}}\\
    \hline
    \textbf{Prediction Change: {\color{red} Real News} $\longrightarrow$ {\color{blue} Fake News}}\\
    \hline
    \end{tabular}
    }
    \vspace*{-17pt}
    \label{tab:realdemotion}
\end{figure}

Circulation of fake news, i.e., false or misleading pieces of information, on social media is not only detrimental to individuals' knowledge but is also creating an erosion of trust in society. Fake news has been promoted with deliberate intention to widen political divides, to undermine citizens' confidence in public figures, and even to create confusion and doubts among communities \cite{hindman2018disinformation}. Hence, any quantity of fake news is intolerable and should be carefully examined and combated \cite{allen2020evaluating}. Due to the high-stakes of fake news detection in practice, therefore, tremendous efforts have been taken to develop fake news detection models that can auto-detect fake news with high accuracies \cite{aldwairi2018detecting,ruchansky2017csi,shu2019defend,cui2019same}. 
Figure \ref{tab:realdemotion} (on top) shows an example of a typical news article posted on the social media channels such as Twitter and Facebook. A fake news detection model then uses different features of the article (e.g., headline and news content) and outputs a prediction on whether such an article is real or fake. Further, recent research has shown that users' engagement (e.g., user comments or replies) on public news channels on which these articles are shared  
become a critical signal to flag questionable news\cite{ruchansky2017csi}. Hence, some of the state-of-the-art (SOTA) fake news detection models \cite{ruchansky2017csi,shu2019defend,cui2019same,qian2018neural} have exploited these user engagement features into their prediction models with great successes. 

Despite the good performances, the majority of SOTA detectors are deep learning based, and thus become vulnerable to the recent advancement in adversarial attacks \cite{papernot2016limitations}.
As suggested by \cite{fakenewsvulnerable}, for instance, a careful manipulation of the title or content of a news article can mislead the SOTA detectors to predict fake news as real news and vice versa. \cite{horne2019robust} also shows that hiding questionable content in an article or replacing the source of fake news to that of real news can also achieve the same effect. However, these existing attack methods suffer from three key limitations: (i) unless an attacker is  also the publisher of fake news, she cannot exercise \textit{post-publish} attacks, i.e., once an article is published, the attacker cannot change its title or content; (ii) an attacker generates adversarial texts either by marginally tampering certain words or characters using pre-defined templates (e.g., ``hello" $\rightarrow$ ``he11o", ``fake" $\rightarrow$ ``f@ke" \cite{li2018textbugger}),  appending short random phrases (e.g., ``zoning tapping fiennes") to the original text \cite{cer2018universal}, or  flipping a vulnerable character or word (e.g. ``opposition" $\rightarrow$ ``oBposition") \cite{ebrahimi2017hotflip}, all of which can be easily detected by a careful examination with naked eyes; and (iii) they largely focus on the vulnerabilities found in the title and content, leaving social responses, i.e., \textit{comments} and \textit{replies}, unexplored.

Since many SOTA neural fake news detectors exploit users' comments to improve fake news detection, this makes them highly vulnerable from attacks via adversarial comments. Figure \ref{tab:realdemotion} shows an example of such an attack. Before the attack, a fake news detector correctly identifies a real article as real. However, using a malicious comment as part of its inputs, the same detector is misled to predict the article as fake instead. 
Compared with manipulating news title or content, an attack by adversarial comments have several advantages: (i) \textit{accessibility:} as it does not require an ownership over the target article, an attacker can easily create a fake user profile and post malicious comments on any social media news posts; (ii) \textit{vulnerability:} it is less vulnerable than attacking via an article's title or content, as the comments written by general users often have a higher tolerance in their writing quality 
(e.g., using more informal language, slang, or abbreviations is acceptable in user comments) compared to that of an article's title or content. This makes any efforts to detect adversarial comments  more challenging. Despite these advantages, to our best knowledge, there exist few studies on  the vulnerability of neural fake news detectors via malicious comments. 

Therefore, in this paper, we formulate a novel problem of adversarial comment generation to fool fake news detectors. Generating adversarial comments is non-trivial because adversarial comments that are misspelled or irrelevant to the news can raise a red flag by a defense system and be filtered out before it has a chance to fool fake news detector. Thus, we are faced with two challenges: (i) \textit{how to  generate adversarial comments that can fool various cutting-edge fake news detectors to predict target class?}; and (ii) \textit{how to simultaneously generate adversarial comments that are realistic and relevant to the article's content;}
In an attempt to solve these challenges, we propose \textbf{\textsc{\mymethod}}, a novel framework that can generate realistic and relevant comments in an end-to-end fashion to attack fake news detection models, that works for both black box and white box attacks. 
The main contributions are:

\begin{itemize}[leftmargin=\dimexpr\parindent-0.2\labelwidth\relax]
    \item This is the first work proposing an attack model against neural fake news detectors, in which adversaries can post malicious comments toward news articles to mislead cutting-edge fake news detectors.
    \item Different from prior adversarial literature, our work generates adversarial texts (e.g., comments, replies) with \textit{high quality} and \textit{relevancy} at the sentence level in an end-to-end fashion (instead of the manipulation at the character or word level).
    \item Our model can fool five top-notch neural fake news detectors to \textit{always} output real news and fake news 94\% and  93.5\% of the time on average. Moreover, our model can mislead black-box classifiers to always output real news 90\% of the time on average.
\end{itemize}
\section{Related Work}
\subsection{Fake News Detection Models}
In terms of computation, the majority of works focus on developing machine learning (ML) based solutions to automatically detect fake news. Feature wise, most models use an article's title, news content, its social responses (e.g., user comments or replies) \cite{shu2019defend}, relationships between subjects and publishers \cite{FAKEDETECTOR} or any combinations of them \cite{ruchansky2017csi}. Specifically, social responses have been widely adopted and proven to be strong predictive features for the accurate detection of fake news \cite{shu2019defend,ruchansky2017csi}. Architecture wise, most detectors use \textit{recurrent neural network (RNN)} \cite{ruchansky2017csi,shu2019defend} or \textit{convolutional neural network (CNN)}  \cite{qian2018neural} to encode either the news content (i.e., article's content or micro-blog posts) or the sequential dependency among social comments and replies. Other complex architecture includes the use of \textit{co-attention} layers \cite{vaswani2017attention} to model the interactions between an article's content and its social comments (e.g., d\textsc{EFEND}~\cite{shu2019defend}) and the adoption of variational auto-encoder to generate synthetic social responses to support early fake news detection (e.g., \textsc{TCNN-URG}~\cite{qian2018neural}).

\subsection{Attacking Fake News Detectors}
Even though there have been several works on general adversarial attacks, very few addressed on the attack and defense of fake news detectors. \cite{fakenewsvulnerable} argues that fake news models purely based on \textit{natural language processing (NLP)} features are vulnerable to attacks caused by small fact distortions in the article's content. Thus, they propose to use a fact-based knowledge graph curated from crowdsourcing to augment a classifier. In a similar effort, \cite{horne2019robust} examines three possible attacks to fake news detectors. They are hiding questionable content, replacing features of fake news by that of real news, and blocking the classifiers to collect more training samples. The majority of proposed attacks leverage an article's title, content, or source. They assume that the attacker has a full ownership over the fake news publication (thus can change title or content). This, however, is {\em not} always the case. In this paper, therefore, we assume a stricter attack scenario where the attacker has no control over the article's source or content, particularly in the case where the attacker is different from the fake news writer. Moreover, we also conjecture that the attacker can be hired to either: (i) promote fake news as real news and (ii) demote real news as fake news to create confusion among the community \cite{hindman2018disinformation}. To achieve this, instead of focusing on attacking an article's content or source, we propose to generate and inject \textbf{new} malicious comments on the article to fool fake news detectors.

\begin{figure*}[t!]
  \centering
  \includegraphics[width=\linewidth]{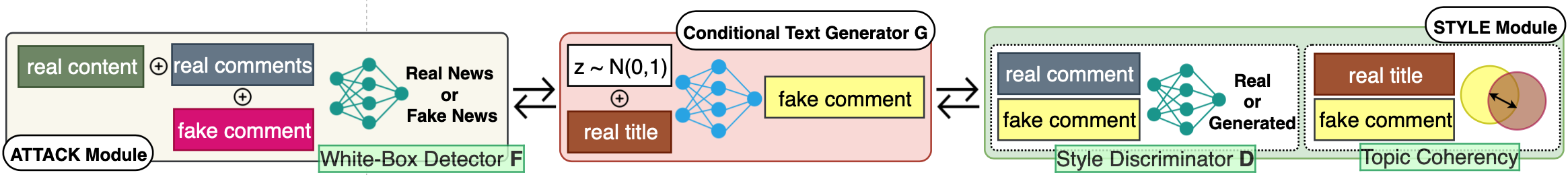}
  \caption{{\mymethod} Architecture.}
  \label{fig:architecture}
  \vspace{3pt}
\end{figure*}

\subsection{Adversarial Text Generation}
Text generation is notoriously a complex problem mainly due to the discrete nature of text. Previous literature in text generation include generating clickbaits \cite{shu2018deep,5sources}, text with sentiment \cite{hu2017toward}, user responses \cite{qian2018neural}, and fake news \cite{zellers2019defending}. Generating text under adversarial setting, i.e., to attack ML classifiers, is more challenging \cite{guo2019ctextgen}. Yet there have been tireless efforts to construct adversarial samples to attack text-based ML models \cite{li2018textbugger,ebrahimi2017hotflip,wallace2019universal}. Most of them focus on making marginal modifications (e.g., addition, removal, replacement, etc.) in character \cite{li2018textbugger,ebrahimi2017hotflip} or word level \cite{wallace2019universal,ebrahimi2017hotflip} of a span of text, either through a set of predefined templates \cite{li2018textbugger} or through a searching mechanism with constraints \cite{wallace2019universal,ebrahimi2017hotflip}. Even though these methods have achieved some degree of success, they are only designed for attacking static features such as the title and content of an article. They are not developed for dynamic sequential input like comments where new text can be added over time. Adversarial text generated by previous methods are usually misspelled ("f@ke" v.s. "fake", "lo ve" v.s. "love") \cite{li2018textbugger}, or distorted from the original context or meaning (e.g., \cite{li2018textbugger}, \cite{ebrahimi2017hotflip}). Hence, these attacks can easily be filter-out by a robust word recognizer (e.g. \cite{pruthi2019combating}) or even by manual visual examination. Because of this, we propose an end-to-end framework to generate stealthy and context-dependent adversarial comments that achieve a high attack performance.

\section{Problem Formulation}\label{sec:attacker}
We propose to attack fake news detectors with three phrases. Phrase I: identifying target articles to attack. Phrase II: generating malicious comments. Phrase III: appending generated comments on the target articles. In this paper, we focus on the \textit{phrase II} of the attack, which is formally defined as follows. Let $\mathit{f}(\cdot)$ be a target neural network fake news classifier. Denote $\mathcal{X} = \{\pmb{x}_i,C_i\}^N_{i=1}$, $\mathcal{Y} = \{y_i\}^N_{i=1}$ as the features of articles and their ground-truth labels (e.g., fake news or real news) of a dataset $\mathcal{D}$ on which $\mathit{f}(\cdot)$ is trained, with $N$ being the total number of articles. Let $\pmb{x}_i^{title}$, $\pmb{x}_i^{content}$, $C_i$ be the title, content, and a list of all comments of $\pmb{x}_i$, respectively. Then, we want to train a generator $G$ such that, given an unseen article $\{\pmb{x},C\} \not\in \mathcal{X}$ and a target prediction label $L^*$, $G$ generates a set of $M$ malicious comments $C^{adv}$ to achieve the following objectives:\\
\vspace{\myspace}

\noindent\textbf{Objective 1:} High quality in writing and relevancy: ${C}^{adv}$ needs to mimic real comments both in writing quality and relevancy to $\pmb{x}$'s content. This will prevent them from being detected by a robust adversarial text defense system (e.g., ~\cite{pruthi2019combating,wang2019learning}). Even though generating realistic comments \cite{shu2018deep} is not the main goal of our paper, it is a necessary condition for successful attacks in practice. \\
\vspace{\myspace}

\noindent\textbf{Objective 2:} Successful attacks: This is the main objective of the attacker. The attacker contaminates a set of an article's existing comments $C$ by \textbf{appending} ${C}^{adv}$ such that $\mathit{f}: \pmb{x}, C^* \mapsto L^*$,
where $C^* \longleftarrow C \oplus {C}^{adv}$ with $\oplus$ denoting concatenating, and $L^*$ is the target prediction label. When $L^*\leftarrow0$, ${C}^{adv}$ ensures that, after posted, an article $\pmb{x}$ will not be detected by $\mathit{f}$ as fake (and not to be removed from the news channels). When $L^*\leftarrow 1$, ${C}^{adv}$ helps demote real news as fake news (and be removed from the news channels). There are two types of attacks: (i) white box and (ii) black box attack. In a white box attack, we assume that the attacker has access to the parameters of $\mathit{f}$. In a black box attack, on the other hand, the $\mathit{f}$'s architecture and parameters are unknown to adversaries. This leads to the next objective below.\\
\vspace{\myspace}

\noindent\textbf{Objective 3:} Transferability: ${C}^{adv}$ needs to be transferable across different fake news detectors. In a black box setting, the attacker uses a surrogate white box fake news classifier $\mathit{f}^*$ to generate ${C}^{adv}$ and transfer ${C}^{adv}$ to attack other black box models $\mathit{f}$. Since fake news detectors are high-stack models, we impose a stricter assumption compared to previous literature (e.g., \cite{li2018textbugger}) where public APIs to target fake news classifiers are inaccessible.
In practice, the training dataset of an unseen black box model will be different from that of a white box model, yet they can be highly overlapped. Since fake news with reliable labels are scarce and usually encouraged to be publicized to educate the general public (e.g., via fact-check sites), fake news defenders have incentives to include those in the training dataset to improve the performance of their detection models. To simplify this, we assume that both white box and black box models share the same training dataset.

\section{Adversarial Comments Generation}

In this paper, we propose \textsc{{\mymethod}}, an \textit{end-to-end}  \textit{\underline{\textbf{Mal}}icious} \textit{\underline{\textbf{Com}}ment Generation Framework}, to attack fake news detection models. Figure \ref{fig:architecture} depicts the {\mymethod} framework. Given an article, {\mymethod} generates a set of malicious comments using a conditional text generator $G$. We train $G$ together with \textsc{Style} and \textsc{Attack} modules. While the \textsc{Style} module gradually improves the writing styles and relevancy of the generated comments, the \textsc{Attack} module ensures to fool the target classifier. 

\subsection{Conditional Comment Generator: \textit{G}}
$G(\pmb{x},\mathbf{z})$ is a conditional sequential text generation model that generates malicious comment $c^*$ by sampling one token at a time, conditioned on (i) previously generated words, (ii) article $\pmb{x}$, and (iii) a random latent variable $\mathbf{z}$. Each token is sequentially sampled according to conditional probability function:
\begin{equation}
\begin{aligned}
p(c^*|\pmb{x};\theta_G) = \prod_{t=1}^{T}p(c^*_t|c^*_{t-1},c^*_{t-2},\dots,c^*_{1};\pmb{x};\mathbf{z})
\end{aligned}
\label{eqn:g_sample}
\end{equation}
where $c^*_t$ is a token sampled at time-step $t$, $T$ is the maximum generated sequence length, and $\theta_{G}$ is the parameters of $G$ to be learned. $G$ can also be considered as a conditional language model, and can be trained using \textit{MLE} with the \textit{teacher-forcing} \cite{lamb2016professor} by maximizing the negative log-likelihood (\textit{NLL}) for all comments conditioned on the respective articles in $\mathcal{X}$. We want to optimize the objective function:
\vspace{-5pt}
\begin{equation}
\min_{\theta_G}\mathcal{L}_G^{MLE} = - \sum_{i=1}^N c_i \log p(c^*_i|\pmb{x}_i;\theta_G)
\label{eqn:mle}
\end{equation}

\subsection{Style Module}
Both writing style and topic coherency are crucial for a successful attack. Due to its high-stake, a fake news detector can be self-guarded by a robust system where misspelled comments or ones that are off-topic from the article's content can be flagged and deleted. To overcome this, we introduce the \textsc{Style} module to fine-tune $G$ such that it generates comments with (i) high quality in writing and (ii) high coherency to an article's content.

First, we utilize the \textit{GAN} \cite{goodfellow2014generative} and employ a comment style discriminator $D$ to co-train with $G$ in an adversarial training schema. We use \textit{Relativistic GAN (RSGAN)} \cite{jolicoeur2018relativistic} instead of standard GAN loss \cite{goodfellow2014generative}. 
In RSGAN, the generator $G$ aims to generate realistic comments to fool a discriminator $D$, while the discriminator $D$ aims to discriminate whether the comment $c$ is more realistic than randomly sampled fake data generated by $G$. 
Specifically, we alternately optimize $D$ and $G$ with the following two objective functions:
\begin{equation}
\begin{aligned}
    \min_{\theta_{G}} \mathcal{L}_{G}^{D} &= -\mathbb{E}_{(x, c) \sim p_D(\mathcal{X}); z \sim p_z}[log(\sigma(D(c) - D(G(\pmb{x},z)))] \\
    \min_{\theta_{D}} \mathcal{L}_{D} &= -\mathbb{E}_{(x, c) \sim p_D(\mathcal{X}); z \sim p_z}[log(\sigma(D(G(\pmb{x},z)) - D(c)))] \\
\end{aligned}
\label{eqn:g_d}
\end{equation}
where $\sigma$ is a $sigmoid$ function, $\theta_{G}$ is the parameters of $G$ and $\theta_{D}$ is the parameters of $D$. By using $D$, we want to generate comments that are free from misspellings while resembling realistic commenting styles. 

Second, to enhance the relevancy between the generated comments and the article, we minimize the mutual information gap between comments generated by $G$ and the article's titles. Specifically, we use \textit{maximum mean discrepancy (MMD)}, which has been shown to be effective in enforcing mutual information. The loss function can be written as:
\begin{equation}
    \begin{aligned}
    \min_{\theta_G}{\mathcal{L}_G^H} &= MMD(\mathcal{X}^{title}, G(\mathcal{X}))\\
     &= \big[\mathbb{E}_{\pmb{x},\pmb{x'} \sim p_{\mathcal{D}}(\mathcal{X})}\mathbf{k}(\pmb{x}^{title},\pmb{x'}^{title})\\ &+ \mathbb{E}_{\pmb{x} \sim p_{\mathcal{D}}(\mathcal{X}); c^*,c^{*'} \sim G(\pmb{x},z)}\mathbf{k}(c^*, c^{*'}) \\
    &- 2\mathbb{E}_{\pmb{x} \sim p_{\mathcal{D}}(\mathcal{X}),c^*\sim G(\pmb{x},z) }\mathbf{k}(\pmb{x}^{title}, c^*)\big]^{\frac{1}{2}}
    \end{aligned}
    \label{eqn:g_h}
\end{equation}
where the $MMD$ compares the distribution $\mathcal{X}^{title}$ of real articles' titles and that of generated comments $G(\mathcal{X})$ by projecting them into Hilbert spaces (RKHS) using a Gaussian kernel $\mathbf{k}$. Intuitively, we want to minimize the information gap between the real titles and the generated comments. Moreover, we use $\pmb{x}^{title}$ (i.e., the title of $\pmb{x}$) instead of $\pmb{x}^{content}$ (i.e., the content of $\pmb{x}$) because: (i) an article's content is usually much longer, hence requiring more computation, (ii) an article's title is a summary of the article's content, and (iii) prior studies (e.g., \cite{userclick}) show that social media users actually rely more on the headlines rather than the actual content for commenting, sharing, and liking.

\subsection{Attack Module}\label{sec:attack_module}
This module guides $G$ to generate comments that can fool the target classifier. In a white box setting, the fake news classifier can be directly used to guide the learning of $G$. In a black box setting, a surrogate fake news classifier can be used to generate and transfer malicious comments to unseen fake news detectors. We denote $\mathit{f}(\pmb{x}_i, \mathbf{C}_i)$ parameterized by $\theta_f$ as the surrogate white box classifier, predicting whether or not $\pmb{x}_i$ is fake news. $f$ can be trained using \textit{binary-cross-entropy loss} over $\mathcal{D}$ as:
\begin{equation*}
\min_{\theta_{f}} \mathcal{L}_{f} = -\frac{1}{N}\sum_{i}^N(\mathbf{y}_i log(\mathit{f}(\pmb{x}_i, C_i))) + (1-\mathbf{y}_i)log(1- \mathit{f}(\pmb{x}_i, C_i))
\label{eqn:f_main}
\end{equation*}

To use this trained model $f$ to guide $G$, we use signals back-propagated from $f$ to force $G$ to generate a new comment $c^*$ such that $\mathit{f}(\pmb{x}_i,C_i^*)$ (where $C_i^* \longleftarrow C_i \oplus\{c^*\}$) outputs a target prediction label $\mathbf{L^*} \in \{0, 1\}$ for the article $\pmb{x}_i$. Specifically, we want to optimize the objective function:
\begin{equation}
\small
\min_{\theta_{G}} \mathcal{L}_{G}^{f(\mathbf{L}^*)} = -\frac{1}{N}\sum_{i}^N(\mathbf{L^*} log(\mathit{f}(\pmb{x}_i, C_i^*))) + (1-\mathbf{L^*})log(1- \mathit{f}(\pmb{x}_i, C_i^*))
\label{eqn:g_f}
\end{equation}
One obvious attack scenario is for an attacker to promote fake news, i.e., to generate comments to mislead the target classifier to classify fake news as real news ($\mathbf{L^*}\longleftarrow0$). Adversely, an attacker might also want to fool the target classifier to classify real news as fake news ($\mathbf{L^*}\longleftarrow1$).


\begin{algorithm}[tb]
    \footnotesize
    \caption{Generating Adversarial Comments Algorithm}
    \label{alg:training}
    \begin{algorithmic}[1]
        \STATE Pre-train $G$ with \textit{teacher-forcing} and MLE using Eq. (\ref{eqn:mle}) with \textit{train} set.
        \STATE Pre-train a surrogate fake news classifier $\mathit{f}$ using Eq. (\ref{eqn:f_main}) with \textit{train} set.
        \REPEAT
            \STATE Training $G$ with $D$ using Eq. (\ref{eqn:g_d}) in mini-batch from \textit{train} set.
            \STATE Training $G$ using Eq. (\ref{eqn:g_h}) in mini-batch from \textit{train} set.
            \STATE Training $G$ with $\mathit{f}$ using Eq. (\ref{eqn:g_f}) in mini-batch from \textit{train} set.
        \UNTIL{convergence}
    \end{algorithmic}
\end{algorithm}

\subsection{Objective Function of {\mymethod}}
At the end, an attacker aims to generate realistic and relevant comments to attack a target fake news classifier by optimizing objective functions as follows.
\begin{equation}
    \min_{\theta_{f}} \mathcal{L}_f;\quad \min_{\theta_{D}} \mathcal{L}_D; \quad \min_{\theta_{G}} (\mathcal{L}_{G}^{MLE} + \mathcal{L}_G^D + \mathcal{L}_G^{H} + \mathcal{L}_{G}^{f(\mathbf{L}^*)})
    \label{eqn:final}
\end{equation}
where each term in Eq. (\ref{eqn:final}) equally contributes to the final loss function. We use Adam \cite{goodfellow2016deep} to optimize the objective functions with a mini-batch training approach. Alg. \ref{alg:training} shows the overall training algorithm. 

\subsection{Implementation Details}\label{sec:implementation}

\emph{Training with Discrete Data:} We need to back-propagate the gradients of the loss in Eq. (\ref{eqn:g_d}, \ref{eqn:g_h}, \ref{eqn:g_f}) through discrete tokens sampled by $G$ from the multinomial distribution $c^*_t$ at each time-step $t$. Since this sampling process is not differentiable, we employ \textit{Gumbell-Softmax} \cite{jang2016categorical} relaxation trick with a $\tau$ parameter (i.e., generation temperature) to overcome this. We refer interested readers on the elaborate discussion of the Gumbell-Softmax technique and the effects of $\tau$ on generation quality and diversity to \cite{nie2018relgan,jang2016categorical}.

\emph{Generation Strategy:}
For each article $\pmb{x}$, we generate new comment $c \longleftarrow G(\pmb{x},z)$ where $z \sim \mathcal{N}(0,1)$. To minimize the risk of being detected by a defender, an attacker desires to select the best set of comments to attack, especially those that are highly relevant to the target article. Hence, for each article $\pmb{x}$, we sample different $z$ to generate different malicious comments and select $c$ that is the most coherent to the article. To measure such the coherency, we derive function $T_k(c, \pmb{x}^{title})$ which will be introduced in Sec. (\ref{sec:evaluation:metrics}).

\emph{Architectures and Parameters Setting:} We employ \textit{Relational Memory Recurrent Network (LMRN)} \cite{nie2018relgan,santoro2018relational} and multi discriminative representations (MDR) \cite{nie2018relgan} as the corner-stone of $G$ and $D$ architecture. We also observe that $D$ with a CNN-based architecture works very well in our case. The LMRN model is adapted from the \textsc{sonnet} model\footnote{\url{https://github.com/deepmind/sonnet}}. The MDR implementation is publicly available \footnote{\url{https://github.com/williamSYSU/TextGAN-PyTorch}}. We will release all datasets, codes, and parameters used in our experiments upon the acceptance of this paper.

\section{Evaluation}

In this section, we evaluate the effectiveness of {\mymethod} and try to answer the following analytical questions (AQs):

\begin{enumerate}[label=\textbf{AQ\arabic*}]
    \item \textbf{Quality, Diversity, and Coherency:} How realistic are the generated comments in terms of their writing styles and as well as coherency to the original articles' contents?\label{aq1}
    \item \textbf{Attack Performance:} How effective are generated comments in attacking white box and black box detectors?\label{aq2}
    \item \textbf{Attack Robust Fake News Detectors:} How effective are generated comments in attacking fake news detectors safe-guarded by a robust comments filtering feature? \label{aq3}
    \item \textbf{Robustness:} How many malicious comments do we need and how early can they effectively attack the detectors?\label{aq4}
\end{enumerate}
\textit{We plan to release all datasets, codes, and parameters used in our experiments (upon the acceptance of this paper).}
\begin{table}[tb]
\centering
\footnotesize
\caption{Dataset Statistics and Details of Target Classifiers and Their Fake News Detection Performance}
\label{tab:notations}
    \begin{tabular}{|l|cccc|}
    \hline
    Dataset & \#articles & \#comments & \#fake & \#real\\
    \hline
    \textsc{GossipCop} & 4,792 & 116,308 & 1,894 & 2,898 \\
    \textsc{Pheme} & 5,476 & 52,612 & 1,980 & 3,486 \\
    \hline
    \hline
    \multirow{2}{*}{Classifier} &  \multicolumn{2}{c|}{\textsc{GossipCop}} & \multicolumn{2}{c|}{\textsc{Pheme}}\\
    \cline{2-5}
    {} & Accuracy & F1 & Accuracy & F1\\
    \hline
    $\mathit{f}_{CNN}$  & 0.74 & 0.74 & 0.77 & 0.77\\
    $\mathit{f}_{RNN}$ & 0.70 & 0.69 & 0.71 & 0.71\\
    \textsc{CSI\textbackslash}t \cite{ruchansky2017csi} & 0.65 & 0.70 & 0.61 & 0.61\\
    \textsc{textCNN} \cite{kim2014convolutional} & 0.68 & 0.68 & 0.76 & 0.76\\
    d\textsc{EFEND} \cite{shu2019defend} & 0.76 & 0.76 & 0.78 & 0.78 \\
    \hline
    \end{tabular}
    \vspace{-15pt}
    \label{tab:dataset_classifiers}
\end{table}

\subsection{Experimental Set-Up}
\subsubsection{Datasets}\label{sec:dataset}
We experiment with two popular public benchmark datasets, i.e., \textbf{\textsc{GossipCop}} \cite{shu2018fakenewsnet} and \textbf{\textsc{Pheme}} \cite{pheme-veracity}. \textsc{GossipCop} is a dataset of fake and real news collected from a fact-checking website, \textit{GossipCop}, whereas \textsc{Pheme} is a dataset of rumors and non-rumors relating to nine different breaking events. These datasets are selected because they include both veracity label and relevant social media discourse content on Twitter\footnote{We exclude another popular dataset, \textsc{PolitiFact}, also from \cite{shu2018fakenewsnet} because it is much smaller and less diverse in terms of topics}. 

\subsubsection{Data Processing and Partitioning}
For each dataset, we first clean all of the comments (e.g., remove mentions, hashtags, URLs, etc.). We also remove non-English comments, and we only select comments that have length from 5 and 20. We split the original dataset into \textit{train} and \textit{test} set with a split ratio of 9:1. Since \textsc{Pheme} dataset does not include articles' contents, we use their titles as alternatives. Table \ref{tab:dataset_classifiers} shows the statistics of the post-processed datasets. We use the \textit{train} set to train both $G$ and target fake news classifiers $f$. All the experiments are done \textit{only} on the \textit{test} set, i.e., we evaluate quality and attack performance of generated comments on {\em unseen} articles and their ground-truth comments.

\subsubsection{Target Classifier}\label{sec:evaluation:classifiers}.
We experiment {\mymethod} with SOTA and representative fake news classifiers, which are summarized in Table \ref{tab:dataset_classifiers}. Note that both datasets are challenging ones as SOTA methods can only achieve 0.76 and 0.78 in F1 using the first 10 comments of each article. These classifiers are selected because they cover a variety of neural architectures to learn representations of each article and its comments, which is eventually input to a \textit{softmax} layer for prediction. In all of the following classifiers, we encode the article's content into a feature vector of $\mathbb{R}^{512}$ by using Universal Sentence Encoder (USE) \cite{cer2018universal}, followed by a \textit{fully-connected-network (FCN)} layer.
\begin{enumerate}[leftmargin=*,label=$\bullet$]
    \item \textbf{$\mathit{\pmb{f}}_{\pmb{CNN}}$}: This classifier uses CNN layers to encode each comment into a vector. Then, it concatenates the average of all encoded comments with the feature vector of the article's content as the article's final representation.
    \label{cnn}
    
    \item \textbf{$\mathit{\pmb{f}}_{\pmb{RNN}}$}: This classifier uses a RNN layer to model the sequential dependency among its comments, output of which is then concatenated with the vectorized form of the article's content as the article's final representation. We utilize \textit{Gated Recurrent Unit (GRU)} as the RNN architecture because it has been widely adopted in previous fake news detection literatures (e.g., \cite{shu2019defend,ruchansky2017csi}). \label{rnn}
    
    \item \textbf{\textsc{textCNN}} \cite{kim2014convolutional}: \textsc{textCNN} uses a CNN architecture to encode the mean of vector representations of all of its comments. The output vector is then concatenated with the article's content vector as the article's final representation.
    
    \item \textbf{\textsc{CSI}\textbackslash t} \cite{ruchansky2017csi}: \textsc{CSI} uses \textit{GRU} to model the sequential dependency of textual features of user comments and the network features among users participating in an article's discourse to detect fake news. Different from $\mathit{f}_{RNN}$, this model does not use an article's content as an input. We use a modified version, denoted as \textsc{CSI}\textbackslash t, that does not use the network features as such information is not available in both datasets.
    
    \item \textbf{d\textsc{EFEND}} \cite{shu2019defend}: Given an article, this algorithm utilizes a co-attention layer between an article's content and comments as input to make a final prediction. 
\end{enumerate}
Other methods are surveyed but not included in the experiments because: (i) overall, their accuracies were reported inferior to those of \textbf{d\textsc{EFEND}}, \textbf{\textsc{CSI}}, (ii) \textsc{SAME}\cite{cui2019same} only uses extracted sentiments of the comments, not the whole text as input, (iii) \textsc{FAKEDETECTOR}\cite{FAKEDETECTOR} mainly uses graph-based features, which is not within our scope of study, and (iv) TCNN-URG\cite{qian2018neural} focuses only on early fake news detection.

    
    

\subsubsection{Compared Attack Methods} We compared {\mymethod} with representative and SOTA adversarial text generators (Table \ref{tab:compare_attacks}).
\begin{enumerate}[leftmargin=*,label=$\bullet$]
    \item \textbf{\textsc{Copycat} Attack}: We created this method as a trivial attack baseline. \textsc{CopyCat} randomly retrieves a comment from a relevant article in the train set which has the target label. We use \textit{USE} to facilitate semantic comparison among articles' contents. 
    
    \item \textbf{\textsc{HotFlip} Attack \cite{ebrahimi2017hotflip}}: This attack finds the most critical word in a sentence and replaces it with a similar one to fool the target classifier. Since \textsc{HotFlip} does not generate a whole sentence but make modifications on an existing one, we first use the comment retrieved by \textsc{CopyCat} as the initial malicious comment. 
    
    \item \textbf{Universal Trigger (\textsc{UniTrigger}) Attack \cite{cer2018universal}}\footnote{ \url{https://github.com/Eric-Wallace/universal-triggers}}: It searches and appends a \textit{fixed} and \textit{universal} phrase to the end of an existing sentence to fool a text classifier. In this case, we want to find an universal topic-dependent prefix to prepend to every comment retrieved by \textsc{CopyCat} to attack. For a fair comparison and to ensure the coherency with the target article's content, we restrict replacement candidates to the top \textit{q=30} words (for \textsc{GossipCop} dataset) and \textit{q=30} words (for \textsc{Pheme} dataset) representing the article's topic. \textit{q} is chosen such that replacement candidates are distinctive enough among different topics. These words and topics are retrieved from a topic modeling function $LDA_k(\cdot)$. 
    
    \item \textbf{\textsc{TextBugger} Attack \cite{li2018textbugger}}: This method generates ``bugs", i.e., carefully crafted tokens, to replace words of a sentence to fool text classifiers. This attack also requires an existing comment to attack. Therefore, we first use \textsc{CopyCat} to retrieve an initial text to attack. Next, we search for ``bugs" using one of the following strategies \textit{insert, delete, swap, substitute-c, substitute-w} as described in \cite{li2018textbugger} to replace one of the words in a comment that achieves the attack goal. 
\end{enumerate}

\begin{table}[t!]
\centering
\caption{Comparison among Attack Methods}
\label{tab:notations}
\begin{tabular}{|l|c|c|c|}
    \hline
    \multirow{2}{*}{\textbf{Method}} & \multirow{1}{*}{\textbf{end-to-end}} & \multirow{1}{*}{\textbf{generalization}} & \multirow{1}{*}{\textbf{level}}\\
    {} & \textbf{generation} & \textbf{via learning} & \textbf{of attack} \\
    \hline
    \textsc{CopyCat} & \Circle & \Circle & sentence\\
    \textsc{HotFlip} & \Circle & \Circle & character/word\\
    \textsc{UniTrigger} & \Circle & \CIRCLE & multi-level\\
    \textsc{TextBugger} & \Circle & \Circle & character/word\\
    {\mymethod} & \CIRCLE & \CIRCLE & sentence\\
    \hline
    \end{tabular}
    \label{tab:compare_attacks}
\end{table}

\begin{table}[t!]
  \caption{Examples of Generated Malicious Comment. Spans in \textbf{\color{purple}purple} and \textit{italics} are retrieved from the train set and carefully crafted. Spans in \textbf{\color{blue}blue} are generated in end-to-end fashion.}
    \begin{tabular}{|l|l|}
    \hline
    \multirow{1}{*}{Title} & \multirow{1}{6cm}{why hollywood won't cast renee zellweger anymore} \\
    \hline
    \multirow{1}{*}{Content} & \multirow{1}{6cm}{so exactly what led renée zellweger, an oscar (...)} \\
    \hline
    \textsc{CopyCat} & \textit{her dad gave her a great smile} \\
    \textsc{+HotFlip} & \textit{her dad \textbf{\color{purple}\sout{gave} got} her a great smile} \\
    \textsc{+UniTrigger} & \textit{\textbf{\color{purple}edit season edit} her dad gave her a great smile} \\
    \textsc{+TextBugger} & \textit{her dad \textbf{\color{purple}\sout{gave} ga ve} her a great smile}\\
    \hline
    \multirow{1}{*}{{\mymethod}} & \multirow{1}{6cm}{\textbf{\color{blue}why do we need to ruin this season}} \\
    \hline
    \end{tabular}
    \vspace{-10pt}
    \label{tab:examples}
\end{table}

\begin{table}[t!]
  \centering
  \caption{Quality, Diversity, Coherency and White Box Attack}
  \begin{tabular}{|l|cccc|}
    \hline
    \multirow{2}{*}{Model} & \multicolumn{4}{c|}{\textbf{\textsc{GossipCop} Dataset}} \\
    \cline{2-5}
    {} & $\uparrow$Quality & $\downarrow$Diversity & $\uparrow$Coherency & $\uparrow$Atk\%\\
    \hline
    \textsc{CopyCat} & 0.650 & - & 0.585 & 0.497 \\
    \textsc{+HotFlip} & 0.618 & - & 0.565 & 0.803\\
    \textsc{+UniTrigger} & 0.545 & - & 0.725 & 0.929\\
    \textsc{+TextBugger} & 0.643 & - & 0.561 & 0.749 \\
    \hline
    \textsc{Malcom\textbackslash Style} & 0.740 & 2.639 & 0.659 & \textbf{0.986}\\
    \textsc{Malcom} & \textbf{0.759} & \textbf{2.520} & \textbf{0.730} & 0.981\\
    \hline
    \hline
    \multirow{2}{*}{Model} & \multicolumn{4}{c|}{\textbf{\textsc{Pheme} Dataset}} \\
    \cline{2-5}
    {} & $\uparrow$Quality & $\downarrow$Diversity & $\uparrow$Coherency & $\uparrow$Atk \\
    \hline
    \textsc{CopyCat} &  0.697 & - & 0.578 & 0.784\\
    \textsc{+HotFlip} & 0.657  & - & 0.530 & 0.958\\
    \textsc{+UniTrigger} &  0.608 & - & 0.595 & 0.951\\
    \textsc{+TextBugger} & 0.617 & - & 0.528 & 0.975\\
    \hline
    \textsc{Malcom\textbackslash Style} & 0.517 & 2.399 & 0.732 & \textbf{1.000}\\
     \textsc{Malcom} &  \textbf{0.776} & \textbf{1.917} & \textbf{0.812} & 0.966\\
    \hline
    \multicolumn{5}{l}{"-": $NLL_{gen}$ cannot be computed for retrieval-based method}\\
    \multicolumn{5}{l}{All experiments are averaged across 3 different runs}
    \end{tabular}
    \vspace{-15pt}
    \label{tab:quality}
\end{table}

\begin{table*}[t!]
  \caption{\textbf{Black Box} Attack Performance on Different Attack Strategies and Target Classifier Architectures (Atk\%)} 
  \centering
  \footnotesize
  \begin{tabular}{|l|c|ccccc||c|ccccc|}
    \hline
    \multirow{2}{*}{Attack/Model} & \multicolumn{6}{c||}{\textbf{\textsc{GossipCop} Dataset}} &  \multicolumn{6}{c|}{\textbf{\textsc{Pheme} Dataset}} \\
    \cline{2-13}
    {} & \textbf{$\mathit{f}^*_{RNN}$} & \multicolumn{1}{c}{\textsc{$\mathit{f}_{RNN}$}} &
    \multicolumn{1}{c}{\textsc{$\mathit{f}_{CNN}$}} & \multicolumn{1}{c}{\textsc{CSI}\textbackslash t} & \multicolumn{1}{c}{\textsc{textCNN}} & \multicolumn{1}{c||}{d\textsc{EFEND}} & \textbf{$\mathit{f}^*_{RNN}$} & \multicolumn{1}{c}{\textsc{$\mathit{f}_{RNN}$}} &
    \multicolumn{1}{c}{\textsc{$\mathit{f}_{CNN}$}} & \multicolumn{1}{c}{\textsc{CSI}\textbackslash t} & \multicolumn{1}{c}{\textsc{textCNN}} & \multicolumn{1}{c|}{d\textsc{EFEND}}\\
    \hline
    \textsc{Baseline} & 0.416 & 0.509 & 0.499 & 0.516 & 0.548 & 0.652  & 0.533 & 0.514 & 0.498 & 0.606 & 0.537 & 0.543\\
    \hline
    \textsc{CopyCat} & 0.497 & 0.689 & 0.688 & 0.670 & 0.774 & 0.802 & 0.784 & 0.783 & 0.766 & 0.821 & 0.716 & 0.644\\
    \textsc{+HotFlip} & 0.803 & 0.813 & 0.765 & 0.820 & 0.838 & 0.866 & 0.958 & 0.850 & 0.845 & 0.879 & 0.811 & 0.711\\
    \textsc{+UniTrigger} & 0.929 & 0.763 & 0.722 & 0.803 & 0.745 & 0.817 & 0.951 & 0.783 & 0.782 & 0.783 & 0.781 & 0.730\\
    \textsc{+TextBugger} & 0.749 & 0.736 & 0.742 & 0.742 & 0.784 & 0.832 & 0.975 & 0.832 & 0.852 & 0.872 & 0.823 & 0.705\\
    \hline
    \textsc{Malcom\textbackslash Style} & \textbf{0.986} & \textbf{0.973} & \underline{0.939} & \underline{0.875} & \textbf{0.888} & \textbf{0.930}  & \textbf{1.000} & \textbf{0.959} & \textbf{0.965} & \underline{0.880} & \textbf{0.963} & \textbf{0.865}\\
    \textsc{Malcom} & \underline{0.981} & \underline{0.963} & \textbf{0.941} & \textbf{0.911} & \underline{0.876} & \underline{0.912} & \underline{0.966} & \underline{0.893} & \underline{0.893} & \textbf{0.888} & \underline{0.889} & \underline{0.760}\\
    \hline
    \multicolumn{12}{l}{(*) indicates white box attacks. 
    All experiments are averaged across 3 different runs. \textsc{Malcom\textbackslash Style}: \textsc{\mymethod} {\em without} the \textsc{Style} module.}
    \end{tabular}
    \vspace{5pt}
    \label{tab:attack}
\end{table*}

\subsubsection{Evaluation Measures}\label{sec:evaluation:metrics}
\begin{enumerate}[leftmargin=\dimexpr\parindent+0.1\labelwidth\relax]
    \item \textbf{Success Attack Rate (Atk\%)}: This quantifies the effectiveness of the attack. For example, a target-real attack on $f$ with Atk\% score of 80\% indicates that an attacker can fool $f$ to predict \textit{real-news} 80\% of the time on all news articles that $F$ should have otherwise predicted correctly.
    
    \item \textbf{Quality and Diversity}: We use \textbf{BLEU} and \textit{negative-log-likelihood loss} \textbf{(NLL\_gen)} scores to evaluate how well generated comments are in terms of both quality and diversity, both of which are widely adopted by previous text generation literature (e.g., \cite{yu2017seqgan,nie2018relgan,guo2019ctextgen}). While \textit{BLEU} scores depict the quality of the generated text compared with an out-of-sample test set of human-written sentences (the higher the better), \textit{NLL\_gen} signals how diverse generated sentences are (the lower the better). 
    
    \item \textbf{Topic Coherency}: We derive a topic coherency score of a set of arbitrary comments $C$ and its respective set of articles $\mathcal{X}$ of size $N$ as follows:
        $T_k(X,C) = \frac{1}{N}\sum_{i=0}^N [1-\cos(LDA_{k}(\pmb{x}_i^{content}), LDA_{k}(c_i))]$,
    where $\cos(\cdot)$ is a \textit{cosine similarity} function. $LDA_{k}(\cdot)$ is a Latent Dirichlet Allocation (LDA) model that returns the distribution of $k$ different topics of a piece of text. We train $LDA_{k}$ on all of the articles' contents using unsupervised learning with hyper-parameter $k$. The larger the score is, the more topic-coherent the comment gets to the article. Because different comments generation algorithms work well with different values of $k$, for a fair comparison, we report the averaged topic coherency across different values of $k$ as the final \textbf{Coherency score}:
        $Coherency = \sum_{k\in \mathcal{K}} T_k(X,C)$.
    We select $k \in \mathcal{K}$ such that the averaged entropy of topic assignment of each article is minimized, i.e., to ensure that the topic assigned to each article is distinctive enough to have meaningful evaluation. 
\end{enumerate}

\subsection{\ref{aq1}. Quality, Diversity and Coherency}
Tables \ref{tab:examples} and \ref{tab:quality} show the examples of the generated comments by all attacks and their evaluation results on the quality, diversity, and topic coherency. {\mymethod} generates comments with high quality in writing and topic coherency. However, we do not discount the quality of human-written comments. The reason why BLEU scores of real comments, i.e., \textsc{CopyCat}, are lower than that of \textsc{\mymethod} is because they use a more diverse vocabulary and hence reduce n-gram matching chances with reference text in the \textit{test} set. The user-study in Sec. \ref{sec:user_study} will later show that it is also not trivial to differentiate between \textsc{\mymethod}-generated and human-written comments even for human users.

Different from all recent attacks, {\mymethod} is an end-to-end generation framework, and can control the trade-off between quality and diversity by adjusting the $\tau$ parameter accordingly (Sec. \ref{sec:implementation}). Thus, by using a less diverse set of words that are highly relevant to an article, {\mymethod} can focus on generating comments with both high quality and coherency. The \textsc{Style} module significantly improves the writing style and boosts the relevancy of generated comments. Without the \textsc{Style} module, we observe that a model trained with only the \textsc{Attack} module will quickly trade in writing style for attack performance, and eventually become stuck in mode-collapse, i.e., when the model outputs only a few words repeatedly. We also observe that the \textsc{Style} module helps strike a balance between topic coherency and attack performance.

\subsection{\ref{aq2}. Attack Performance}
In this section, we evaluate Atk\% of all attack methods under the most optimistic scenario where the target article is just published and contains no comments. We will evaluate their attack robustness under other scenarios in later sections. 

\subsubsection{White Box Attack}
 We experiment a white box attack with $\mathit{f}_{RNN}$ target classifier. \textit{RNN} architecture is selected as the white box attack due to its prevalent adoption in various fake news and rumors detection models. Table \ref{tab:attack} describes white box attack in the first column of each dataset. We can observe that {\mymethod} is very effective at attacking white box models (98\% Atk\% and 97\% Atk\% in \textsc{GossipCop} and \textsc{Pheme} dataset). Especially, \textsc{\mymethod\textbackslash Style} is able to achieve near perfect Atk\% scores. Comparable to {\mymethod} are \textsc{UniTrigger} and \textsc{TextBugger}. While other attacks such as \textsc{TextBugger} only performs well in one dataset, {\mymethod} and \textsc{UniTrigger} perform consistently across the two datasets with a very diverse writing styles. This is thanks to the ``learning" process that helps them to generate malicious comments from not only a single but a set of training instances. On the contrary, \textsc{TextBugger} for example, only exploits a specific pre-defined weakness of the target classifier (e.g. the vulnerability where misspellings are usually encoded as unknown tokens \cite{li2018textbugger}) and requires no further learning.


\begin{figure}[tb]
  \centering
  \vskip -1em
  \hspace{-0.5cm}
  \subfloat{
    \includegraphics[width=0.5\textwidth]{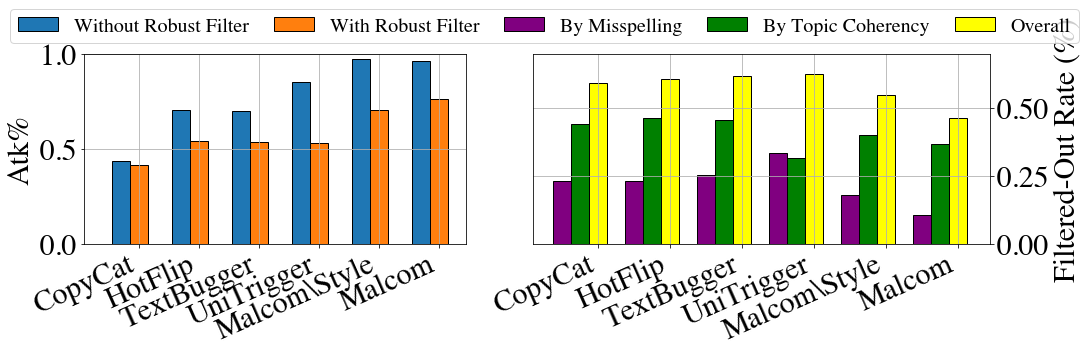}
  }
  \\
  \vspace{-10pt}
  \subfloat{
    \includegraphics[width=0.5\textwidth]{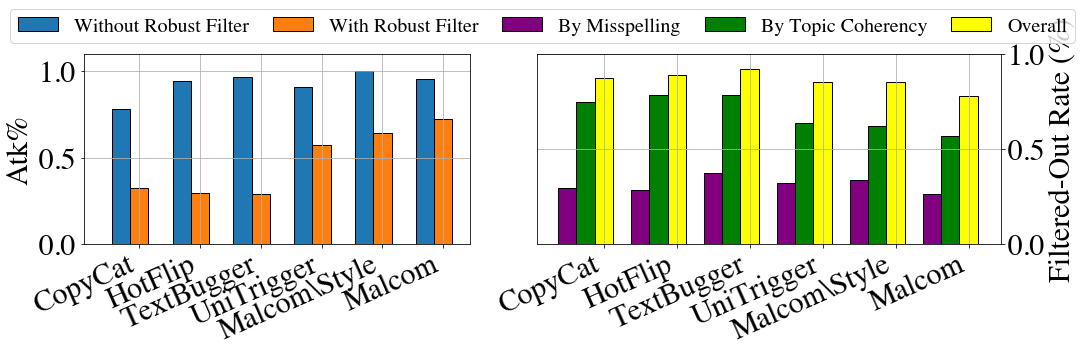}
  }
  \caption{Attack Robust Fake News Detector. Top: \textsc{GossipCop} Dataset. Bottom: \textsc{Pheme} Dataset}
  \label{fig:robust_detector}
  \vspace{-20pt}
\end{figure}

\subsubsection{Black Box Attack}
Let's use a surrogate $\mathit{f}_{RNN}$ model as a proxy target classifier to generate malicious comments to attack black box models. We test their transferability to black box attacks to five unseen fake news classifiers described in Sec. \ref{sec:evaluation:classifiers}. Table \ref{tab:attack} shows that comments generated by \textsc{\mymethod} does not only perform well on white box but also on black box attacks, achieving the best transferability across all types of black box models. Our method is especially able to attack well-known models such as \textsc{CSI\textbackslash}t and d\textsc{EFEND} with an average of 91\% and 85\% of Atk\% in \textsc{GossipCop} and \textsc{Pheme} dataset. However, other strong white box attacks such as \textsc{UniTrigger}, \textsc{TextBugger} and \textsc{HotFlip} witness a significant drop in Atk\% with black box target classifiers. Particularly, \textsc{UniTrigger} experiences the worst transferability, with its Atk\% drops from an average of 94\% to merely over 77\% across all models in both datasets. On the other hand, {\mymethod} performs as much as 90\% Atk\% across all black box evaluations. This shows that our method generalizes well not only on fake news detector trained with different set of inputs (e.g., with or without title), but also with different modeling variants (e.g., with or without modeling dependency among comments) and architectures (\textit{RNN}, \textit{CNN}, \textit{Attention}).

\begin{figure*}[t!]
  \centering
  \hspace*{-1.3cm}
  \includegraphics[width=1.05\linewidth]{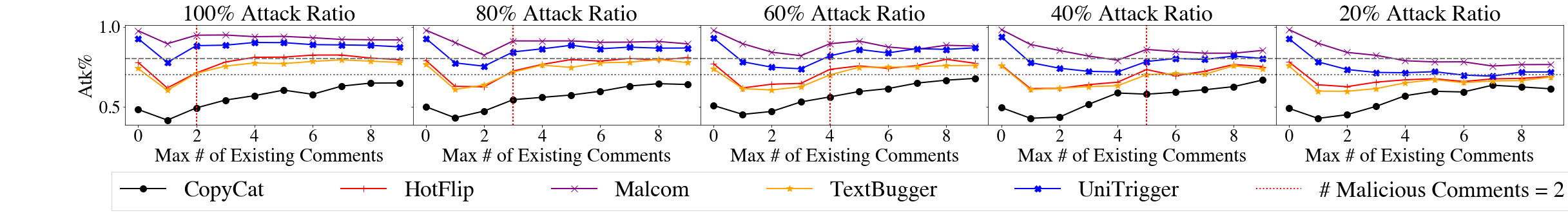}
  \caption{Robustness of Inter-Attacks: White Box Setting on \textsc{GossipCop} Dataset}
  \label{fig:inter_robust_whitebox}
\end{figure*}

\subsection{\ref{aq3}. Attack Robust Fake News Detection}\label{sec:robust}

This section evaluates attack performance of all methods under a \textit{post training defense}. This defense comes after the target classifier has already been trained. Before prediction, we use a robust word recognizer called \textsc{ScRNN}\footnote{\url{https://github.com/danishpruthi/Adversarial-Misspellings}} \cite{pruthi2019combating} to measure the number of misspellings and detect manipulations in the comments. We also use the \textit{Coherency} (Sec. \ref{sec:evaluation:metrics}) to measure the topic relevancy and set a topic coherency threshold between comments and the target article to filter-out irrelevant comments. We remove any comment that either has more than one suspicious word or have \textit{Coherency} lower than that of the article's title with an allowance margin of $0.05$. This defense system is selected because it does not make any assumption on any specific attack methods, hence it is both general and practical.
We measure both Atk\% and the \textit{filter-out rate}, i.e., the percentage of comments that are removed by the defense system, for all of the attack methods.
Figure \ref{fig:robust_detector} shows that our method achieves the best Atk\% even under a rigorous defense in both datasets. While Atk\% of \textsc{HotFlip} and \textsc{TextBugger} drop significantly (around $\downarrow$66\% and $\downarrow$68\%) under the defense, that of \textsc{\mymethod\textbackslash Style} decreases to only 0.64\% from a nearly perfect Atk\%. This confirms that the \textsc{Style} module is crucial for generating stealthy comments. Figure \ref{fig:robust_detector} also shows that {\mymethod} is the best to bypass the defense system, achieving better, i.e., lower, \textit{filter-out rate} in terms of misspellings compared with real comments retrieved by \textsc{CopyCat} method. This is because {\mymethod} emphasizes writing quality over diversity to be more stealthy under a robust defense algorithm. Moreover, around 1/3 of real comments selected by \textsc{CopyCat} are filtered-out by \textsc{ScRNN}. This confirms that real comments written on social media are messy and not always free from grammatical errors.


\begin{figure}[tb]
  \centering
  \vspace{-5pt}
  \hspace{-0.8cm}
  \subfloat{
    \includegraphics[width=0.5\textwidth]{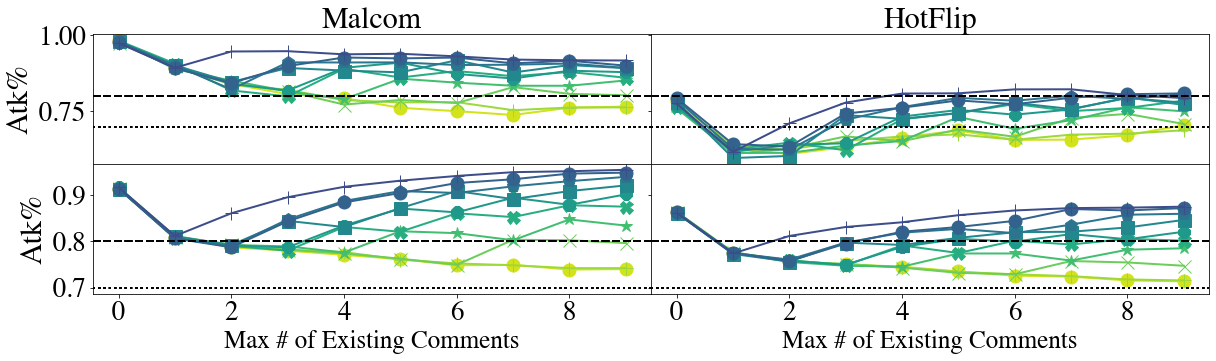}
  }
  \\
  \vspace{-10pt}
  \hspace{-0.8cm}
  \subfloat{
    \includegraphics[width=0.5\textwidth]{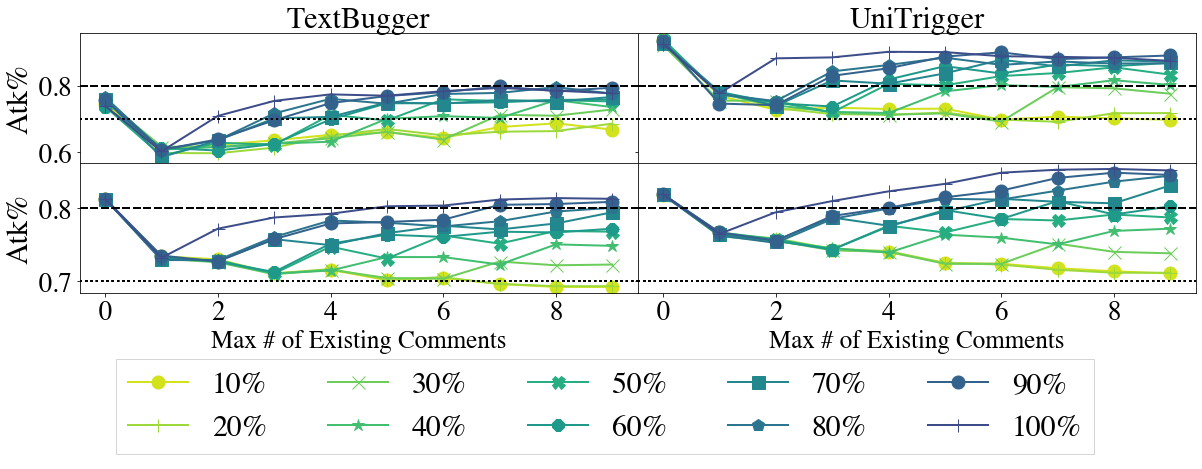}
  }
  \caption{Robustness of Intra-Attacks: White Box Setting (First Row) \& Black Box (Second-Row) on \textsc{GossipCop} Dataset.}
  \label{fig:intra_robust}
  \vspace{-15pt}
\end{figure}

\subsection{\ref{aq4}. Robustness}\label{sec:robustness}
There is always a trade-off between the \# of comments collected for prediction and how early to detect fake news. This section evaluates Atk\% w.r.t different \# of existing comments for early fake news detection. We evaluate on \textsc{GossipCop} dataset as an example. We assume that the target classifier only has access to a few existing comments, a maximum of 20 in this case. First, we define the \textit{attack ratio} as the ratio of \# of malicious comments over \# of existing comments. Next, we evaluate the robustness among all attack methods in two categories, namely \textit{Inter- and Intra-Comparison}.

\textit{Inter-Comparison: How attack methods perform differently with the same attack ratio?} Figure \ref{fig:inter_robust_whitebox} shows that {\mymethod} outperforms other baselines under all attack ratios. Moreover, our method consistently maintains Atk\% of at least 80\% under a different \# of existing comments with an attack ratio as low as 40\%. On the contrary, to achieve the same performance, \textsc{UniTrigger}, the second-best attack in terms of robustness, would require a 100\% attack ratio. 
\textit{Intra-Comparison: For each method, how many malicious comments are needed for an effective attack performance under different \# of existing comments?} Figure \ref{fig:intra_robust} shows the results on white box and black box attacks. Under the white box attack, all methods display more or less the same performance with more than 1 malicious comment. However, the black box setting observes more variance across different attack ratios. Specifically, {\mymethod}'s performance continuously improves as the \# of existing comments increases under any attack ratios $\geq$ 40\%. On the contrary, existing attacks show little improvement even with an increasing \# of malicious comments.

\begin{table}[t!]
    \centering
    \vspace{5pt}
    \caption{Results of User-Study on Generation Quality}
    \begin{tabular}{|c|c|c|c|c|}
    \hline
      Hypothesis & z-score & p-value & Accuracy & \#response \\
      \hline
      $\mathcal{H}_1$ & 1.4940 & 0.0676 & 0.6087 & 46\\
      $\mathcal{H}_2$ & 1.2189 & 0.1114 & 0.5818 & 56\\
      $\mathcal{H}_3$ & 0.9122 & 0.1808 & 0.5416 & 120\\
      \hline
    \end{tabular}
    \vspace{-10pt}
    \label{tab:user_study}
\end{table}

\section{Discussion}
\subsection{Prevent Malicious Comments with Human Support}\label{sec:user_study}
We examine whether malicious comments generated by \textsc{\mymethod} can be easily flagged by human, i.e., the Turing Test. We use Amazon Mechanical Turk (AMT) to recruit over 100 users to distinguish comments generated by \textsc{\mymethod} (machine-generated) and human. We examine the following alternative hypothesises using one-tailed statistical testing.
\begin{enumerate}[leftmargin=\dimexpr\parindent+0.1\labelwidth\relax]
    \item $\mathcal{H}_1$: Given a comment, the users can correctly detect if the comment is generated by machine (not by human).
    \item $\mathcal{H}_2$: Given a comment, the users can correctly detect if the comment is generated by human (not by machine).
    \item $\mathcal{H}_3$: Given a machine-generated and a human-written comment, the users can correctly identify the machine-generated.
\end{enumerate}
For quality assurance, we recruit only the users with 95\% approval rate, randomly swap the choices and discard responses taking less than 30 seconds. We test on comments generated for 187 unseen and unique articles in the \textsc{Pheme} dataset's \textit{test} set. Table \ref{tab:user_study} shows that we \textit{fail to reject} the null-hypothesises of both $\mathcal{H}_1, \mathcal{H}_2$ and $\mathcal{H}_3$ (p-value $>$ 0.05). While comments generated by \textsc{\mymethod} is not perfectly stealthy (accuracy of $\mathcal{H}_1 > 0.5$), it is still very challenging to distinguish between human-written and \textsc{\mymethod}-generated comments. In fact, human-written comments on social media are usually messy, which lead  users to be unable to distinguish between machine-generated ones (accuracy of $\mathcal{H}_2 < 0.6$). Thus, even if human are employed to filter out suspicious or auto-generated comments with a mean accuracy of 60\% ($\mathcal{H}_1$), \textsc{\mymethod} can still effectively achieve over 80\%  Atk\% on average with a remaining 40\% of the malicious comments (see Sec. \ref{sec:robustness}). Hence, we need to equip human workers with intensive training to better identify malicious comments. Nevertheless, this can be labor intensive and costly due to a large amount of comments published everyday. 

\subsection{Prevent Malicious Comments with Machine Support}
One advantage of the defense system introduced in Sec. \ref{sec:robust} is that filtering out comments based on misspellings and topic coherency does not make any assumption on any specific attack methods, hence it is both general and practical. In the most optimistic scenario where we expects {\em only} attacks from \textsc{\mymethod}, we can train a ML model to detect \textsc{\mymethod}-generated comments. We use LIWC \cite{pennebaker2015development} dictionary to extract 91 psycholinguistics features and use them to train a \textit{Random Forest} classifier to differentiate between human-written, i.e., \textsc{CopyCat}, and \textsc{\mymethod}-generated comments based on their linguistic patterns. The 5-fold cross-validation accuracy is $0.68 (+/- 0.1)$, which means around 70\% and 30\% of \textsc{\mymethod}-generated and human-written comments will be flagged and removed by the classifier. From Figure \ref{fig:intra_robust}, if the initial \textit{attack ratio} of 100\%, one can then effectively put an upper-bound of around 80\% Atk\% rate on \textsc{\mymethod} (new \textit{attack ratio} of 40\%). 

Other ways to defend against {\mymethod} is to only allow users with verified identifications to publish or engage in discussion threads, or to utilize a fake account detection system (e.g., \cite{yuan2019detecting}) to weed out suspicious user accounts. We can also exercise adversarial learning \cite{goodfellow2014explaining} and train a target fake news classifier together with malicious comments generated by potential attack methods. Social platforms should also develop their own proprietary fake news training dataset and rely less on public datasets or fact-checking resources such as \textit{GossipCop} and \textit{PolitiFact}. While this may adversely limit the pool of training instances, it will help raise the bar for potential attacks.

\subsection{Real News Demotion Attack}
An attacker can be paid to either promote fake news or demote real news. By demoting real news, in particular, not only can the attacker cause great distrust among communities, but the attacker can also undermine the credibility of news organizations and public figures who publish and share the news. In fact, {\mymethod} can also facilitate such an attack, i.e., to fool fake news detectors to classify real news as fake news, by simply specifying the target label $L^* \longleftarrow 1$ (Sec. \ref{sec:attack_module}). Our experiments show that {\mymethod} can achieve a real news demotion white box attack with Atk\% of around 92\% and 95\% in \textsc{GossipCop} and \textsc{Pheme} datasets. Figure \ref{tab:realdemotion} shows an example of a real news demotion attack. The real news article was posted by \textit{The Guardian}, a reputable news source, on Twitter on June 4, 2018. The article is first correctly predicted as real news by a \textit{RNN}-based fake news classifier. However, the attacker can post a malicious yet realistic-looking comment ``{\it he's a conservative from a few months ago}" to successfully fool the classifier to predict the article as fake instead.


\begin{table}[t!]
  \centering
  \caption{Ablation Test}
  \begin{tabular}{|l|cccc|}
    \hline
    \multirow{2}{*}{Models} & \multicolumn{4}{c|}{\textbf{\textsc{GossipCop} Dataset}} \\
    \cline{2-5}
    {} & $\uparrow$Quality & $\downarrow$Diversity & $\uparrow$Coherency & $\uparrow$Atk\% \\
    \hline
     \textsc{\textbackslash Style\textbackslash Attack} & 0.645 & \textbf{1.664} & 0.727 & 0.392 \\
     \textsc{\textbackslash Attack} & \underline{0.741} &  \underline{2.032} & \textbf{0.769} & 0.434 \\
     \textsc{\textbackslash Style} & 0.740 & 2.639 & 0.659 & \textbf{0.986}\\
     \textsc{\mymethod} & \textbf{0.759} & 2.520 & \underline{0.730} & \underline{0.981} \\
    \hline
    \hline
    \multirow{2}{*}{Models} & \multicolumn{4}{c|}{\textbf{\textsc{Pheme} Dataset}} \\
    \cline{2-5}
    {} & $\uparrow$Quality & $\downarrow$Diversity & $\uparrow$Coherency & $\uparrow$Atk\% \\
    \hline
     \textsc{\textbackslash Style\textbackslash Attack} & 0.759 & \textbf{1.273} & \underline{0.845} & 0.519\\
     \textsc{\textbackslash Attack} & \underline{0.741} & \textbf{0.786} & \underline{1.431} & \textbf{0.850}\\
     \textsc{\textbackslash Style} & 0.517 & 2.399 & 0.732 & \textbf{1.000}\\
     \textsc{{\mymethod}} & \underline{0.776} & 1.917 & \textbf{0.812} & \underline{0.966}\\
    \hline
     \multicolumn{5}{l}{\textsc{\textbackslash Style}, \textsc{\textbackslash Attack}: \textsc{\mymethod} with the \textsc{Style}, \textsc{Attack} module removed.}
    \end{tabular}
    \vspace{-12pt}
    \label{tab:ablation}
\end{table}

\subsection{Ablation Test}
This section carries out ablation test to show the effectiveness of \textsc{Style} and \textsc{Attack} component of {\mymethod}. Specifically, we valuate the quality, diversity, coherency and as well as white box attack performance of different variants of {\mymethod}. Figure \ref{tab:ablation} demonstrates that \textsc{Style} module enhances the writing quality and coherency by large margin from the model without \textsc{Style} module. Especially, \textsc{Style} module is crucial in improving topic coherency score, which then makes generated comments more stealthy under robust fake news defense system (Sec. \ref{sec:robust}). Figure \ref{tab:ablation} also shows that \textsc{Attack} module is critical in improving attack performance. 
While \textsc{Style} and \textsc{Attack} each trades off quality, coherency with attack success rate and vice versa, the full {\mymethod} achieves a balanced performance between a good writing style and high attack success. This makes our framework both powerful and practical.

\subsection{Baselines' Dependency on \textsc{CopyCat}}
Compared to {\mymethod}, one disadvantage of \textsc{HotFlip}, \textsc{UniTrigger} and \textsc{TextBugger} is that they all require an initial comment to manipulate. In theory, manually crafting an initial comment is feasible, yet demands a great labor cost. In practice, an attacker can directly use the target's title or an existing comment as the initial comment to begin the attack. Instead, in this paper, we use \textsc{CopyCat} to retrieve the initial comment. \textsc{CopyCat} considers both the topic of the target article and the target label into consideration. Hence, it can help complement other baseline attacks in terms of both Atk\% and topic coherency. Our experiments show that attacks using comments retrieved by \textsc{CopyCat} achieve much better averaged Atk\% across both white box and black box attacks (89\% Atk\%), compared to the ones using the title (75\% Atk\%) or a single existing comment (78\% Atk\%) of the target article. This further justifies the use of \textsc{CopyCat} together with other baseline attacks in our experiments.

\section{Limitations and Future Work}
In this work, we assume that the attacker and the model provider share the same training dataset. In practice, their training datasets might be overlapped but not exactly the same. Moreover, whether or not comments generated using one sub-domain (e.g., political fake news) can be transferable to another (e.g., health fake news) is also out of scope of this paper. Hence, we leave the investigation on the proposed attack's transferability across different datasets for future work. Moreover, we also plan to extend our method to attack graph-based fake news detectors (e.g., \cite{qian2018neural}), and evaluate our model with other defense mechanisms such as adversarial learning, i.e., to train the target fake news classifier with both real and malicious comments to make it more robust. We also want to exploit similar attack strategy in areas that utilize sequential dependency among text using ML such as fake reviews detection.

\section{Conclusion}
To our best knowledge, this paper is the first attempt to attack existing neural fake news detectors via malicious comments. Our method does not require adversaries to have an ownership over the target article, hence becomes a practical attack. We also introduce {\mymethod}, an end-to-end malicious comments generation framework that can generate realistic and relevant adversarial comments to fool five of most popular neural fake news detectors to predict fake news as real news with attack success rates of 94\% and 90\% for a white box and black box settings. Not only achieving significantly better attack performances than other baselines, {\mymethod} is shown to be more robust even under the condition when a rigorous defense system works against malicious comments. We also show that {\mymethod} is capable of not only promoting fake news but also demoting real news. Due to the high-stakes of detecting fake news in practice, in future, we hope that this work will attract more attention from the community towards developing fake news detection models that are  accurate yet resilient against potential attacks.

\bibliographystyle{IEEEtrans}
\bibliography{IEEEabrv}

\end{document}